\pdfoutput=1
\documentclass[sigconf]{acmart}
\AtBeginDocument{%
  \providecommand\BibTeX{{%
    \normalfont B\kern-0.5em{\scshape i\kern-0.25em b}\kern-0.8em\TeX}}}
    
\renewcommand\footnotetextcopyrightpermission[1]{} %

\usepackage[utf8]{inputenc} %
\usepackage[T1]{fontenc}    %
\usepackage{hyperref}       %
\usepackage{url}            %
\usepackage{booktabs}       %
\usepackage{amsfonts}       %
\usepackage{nicefrac}       %
\usepackage{microtype}      %
\usepackage{graphicx}
\usepackage{subcaption}
\usepackage{booktabs}
\usepackage{url}
\usepackage{filecontents}
\usepackage[Export]{adjustbox}

\begin{document}

\title{AutoML for Contextual Bandits }

\author{Praneet Dutta, Joe Cheuk, Jonathan S Kim, Massimo Mascaro}
\authornote{All authors contributed equally to this research.}
\email{{praneetdutta, joecheuk, jonathanskim, massy} @google.com}

\affiliation{%
  \institution{Google Inc.}
  \streetaddress{1600 Amphitheatre Avenue}
  \city{Mountain View}
  \state{California}
  \postcode{94043}
}

\settopmatter{printacmref=false}
\renewcommand{\shortauthors}{Dutta, et al.}

\begin{abstract}
  Contextual Bandits is one of the widely popular techniques used in applications such as personalization, recommendation systems, mobile health, causal marketing etc \cite{DBLP:journals/corr/abs-1810-01859}. As a dynamic approach, it can be more efficient than standard A/B testing in minimizing regret. We propose an end to end automated meta-learning pipeline to approximate the optimal Q function for contextual bandits problems. We see that our model is able to perform much better than random exploration, being more regret efficient and able to converge with a limited number of samples, while remaining very general and easy to use due to the meta-learning approach. We used a linearly annealed e-greedy exploration policy to define the exploration vs exploitation schedule. We tested the system on a synthetic environment to characterize it fully and we evaluated it on some open source datasets ~\cite{a-contextual-bandit-bake-off} to benchmark against prior work. We see that our model outperforms or performs comparatively to other models while requiring no tuning nor feature engineering. 
\end{abstract}

\keywords{Meta-Learning, AutoML, Contextual Bandits, Regret, Online Learning}

\maketitle
\pagestyle{plain}

\section{Introduction}

Contextual Bandits is a class of dynamic algorithms which can be used to learn efficiently targeting strategies. It is an extension of the multi-armed bandit problem ~\cite{DBLP:journals/corr/abs-1904-07272}, generalizing it with the concept of a context.
Given a  sampled context, the goal of the learning algorithm is to pick an action  which maximizes the reward defined by the environment dynamics. We assume all the actions to be uncorrelated between them (e.g. each state-action-reward is a separate episode). 
In a bandit problem, we can only observe the outcome of an action that has been selected for a given state. The goal of a Bandit formulation is to minimize "regret" - the difference between the cumulative reward from the  optimal policy and the trained agent cumulative sum of rewards. [Eqn \ref{regret_eqn}]

In this work we prove that it is possible to build efficient contextual bandit system by using an off the shelf meta-learning product (Google Cloud AutoML Tables) to learn policies without the need of any algorithmic coding or feature engineering. AutoML is on a high level similar to the Neural Architecture Search (NAS) proposed by \citep{DBLP:journals/corr/ZophL16} which uses an auto-regressive controller to generate architectural hyper-parameters of a Neural Network. Rather than experimenting and hand crafting the best hierarchical representation of deep learning layers, these are learned automatically by the system~\citep{NAS:Survey}. 

We aim to use this meta-learning approach to approximate the Q-function for a contextual bandit, e.g. the expected reward for a given action in a given state:
\begin{equation}
 Q(s,a) = E[r|s,a]
\end{equation}
where r is the reward, s the state and a the action.
Armed with an approximated Q-function generated by AutoML automatically, we can create easily an exploratory policy using the $\epsilon$-greedy exploration schedule. The Q-function, and thus the policy, is updated periodically as new batches of data are accumulated. Being a function of the state space, the Q-function will be able to generalize across it.
Note that we use here the  Reinforcement Learning term "Q-function" even if the system we study is purely a contextual bandit one. We believe the notation and this work can be extended to some multiaction problems by replacing the immediate reward with a long term discounted reward to implement a simple form of Q-Learning. We plan to address this idea in a  future work.

Furthermore, while we limit ourselves to the e-greedy exploration schedule in this work, this work can potentially be applied to other exploration strategies as well. The goal here is to show how using an off-the-shelf AutoML product we can get functioning bandits systems with minimal tuning, thus we focused on the simplest exploration scheme.
\section{Related Work}\label{Related Work}

Our work is inspired by that of Li et al. \citep{DBLP:journals/corr/abs-1003-0146} who proposed the use of a contextual bandit based algorithm which can be evaluated offline. Their model achieved state of the art results on the Yahoo! Front Page Today Module Dataset. Langford et al. \citep{Langford:2007:EAC:2981562.2981665}  proposed a  greedy exploration model which requires no knowledge of the time horizon and were able to determine an upper bound on the regret for their formulation. 

Agarwal et al. \citep{DBLP:journals/corr/AgarwalHKLLS14} utilized an approach which guarantees an upper bound on the number of calls to an Oracle to get the statistically optimal regret guarantee. However, they evaluated their model on a non-public dataset which we could not use against our technique. 

There has been prior work in Meta-learning. Sharaz and Daum{\'{e}} \citep{DBLP:journals/corr/abs-1901-08159}  proposed the use of an Imitation Learning based approach, "M\^EL\'EE" algorithm  based on AggreVaTe. It provides the utility of moving away from hand engineering model architectures.

\section{Mathematical Foundation}\label{Mathematical Foundation}
\subsection{Bandit Formulation}

The Set up for a Contextual Bandit problem is that an agent observes repeatedly a context $s_t$, perform an action $a_t$ and receives a reward $r_t$ that depends (typically stochastically) on both $s_t$ and $a_t$ from the environment ($r_t \sim P(s_t,a_t)$ that from now on we'll simplify as $r_t(s_t,a_t)$ to keep the notation slim, implicitly intending it's a stochastic variable).

The goal is to optimize the cumulative reward across a given sequence of episodes.

\begin{equation}
\max\sum_{t=1}^{T}r_{t}(s_{t},a_{t}) 
\end{equation}
where in the bandit problem we assume that the states at different times are all independent of each other and from the action taken in previous episodes.

A common measure of evaluating the performance of a Contextual Bandit algorithm is to estimate the net "regret". Regret can be framed as the difference between our model's cumulative reward over time and the sequence of actions taken by the most optimal policy over the same period. The goal is to minimize this cost function as quickly as possible in a given period.

\begin{equation}
R_\pi(T) = \max_{\tilde{\pi} \in \Pi}\sum_{i=1}^{i=T}E_{\tilde{\pi}} [r_{t}(s_t,a_{t})] - \sum_{i=1}^{i=T}E_\pi [r_{t}(s_t,a_{t})] \label{regret_eqn}
\end{equation}

\subsection{Meta-Learning Set Up}\label{Meta-Learning}
Google AutoML is inspired by the work on Automatic Architecture Selection \cite{DBLP:journals/corr/ZophL16}. AutoML can help to optimize models that predict the expected reward (payoff) of a given action in a given context.

At a high level, a vanilla AutoML table implementation consists of the following steps in the pipeline. The core pillar of this work is that models are built with the available out of the box tools and with no hand crafted feature engineering or tuning \cite{GoogleAIBlog}.

Under the hood, a multistage Tensorflow Pipeline is automatically instantiated consisting of :-

\begin{itemize}
    \item Automated Feature Engineering is applied to the raw input data.
    \item  Architecture Search to compute the best architecture/(s) for our bandits formulation task, e.g. to find the best predictor model for the expected reward of each episode.
    \item Hyper-parameter Tuning through search
    \item Model Selection. Models which have achieved promising results so far are passed onto the next stage
    \item Model Tuning and Ensembling
\end{itemize}

\section{Datasets}
As stated we ran experiments with multiple data sets to test the efficacy of our meta-learning approach on a variety of different tasks. 

\subsection{Synthetic Dataset} \label{Synthetic Dataset}

\begin{figure}[h!]
    \centering
    \begin{subfigure}[b]{0.9\linewidth}
    \includegraphics[width=\linewidth, height=6cm]{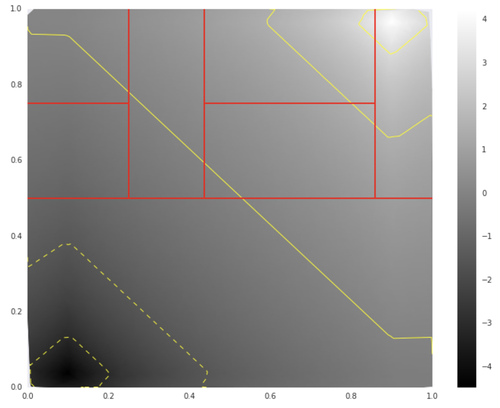}
    \caption{2 Dimension State space with 2 Factors }
    \label{fig:my_label}
    \end{subfigure}
    \begin{subfigure}[b]{0.9\linewidth}
    \includegraphics[width=\linewidth, height=6cm]{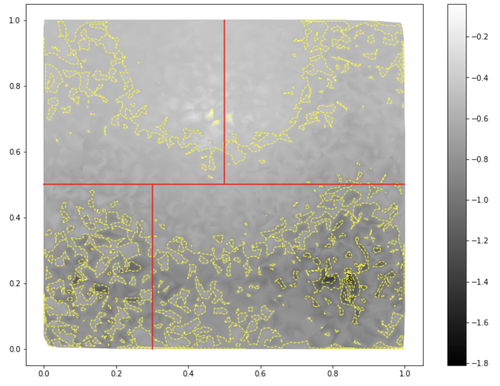}
    \caption{5 Dimensional State Space with 5 factors}
    \label{fig:my_label}
    \end{subfigure}
    \caption{An example of the synthetic environment used for our task. We display the average reward probability across 2 dimensions of the context space. The first example is very simple and represent a case with 2 dimensions in total and only 2 Gaussian factors using to create the distribution. The second case is more complex with the total dimensionality of the state space being 5 (only 2 dimensions are projected here) and 5 Gaussian factors.}
    \label{fig:Synthetic_Dataset}
\end{figure}

We initially created  our own synthetic dataset simulation to provide an experimental platform to test our early results. The simulation sampled an underlying multidimensional and multifactor probability model for which we could tune the complexity artificially. This allowed us to experiment in a controlled environment with state space, distribution as well as the number and type of actions. See fig. \ref{fig:Synthetic_Dataset}.

\subsection{Public Dataset}\label{Public Dataset}

In addition to the synthetic environment, we also compared our model's performance across four public datasets to benchmark the contribution. Some of these datasets have been used in popular work in this field. ~\cite{a-contextual-bandit-bake-off}: 

\begin{itemize}
	\item A scientific dataset intended to simulate gamma particles collected in a atmospheric Cherenkov telescope. The dataset was taken from Chilingarian et al. \citep{Gamma-Telescope-Data-Set}. 
	\item A chess dataset which contains a list of features describing the board set up, and then a class denoting whether or not the white can win from that position, from Dua et al. \citep{Dua:2019}. 
	\item A dataset on forest covertype, containing cartographic information of the forest and also a classification denoting whether the forest coverage consists of primarily Spruce Firs or Lodgepole Pines. The dataset was taken from Blackard and Anderson's \citep{Covertype-Data-Set} work.
	
	\item A dataset that held information about a game called Dou Shou Qi, and a dataset describing the different states of the game board. The classification denoted which player won the game, showcased in Rijin and Vis  \citep{Dou-Shou-Qi}.
\end{itemize}

Each of these real world data sets were partitioned into blocks of 500 to 1000 episodes in size. During each of these blocks we run the current policy and we record all the triples (contexts, action, reward) corresponding to each trial. At the end of the block we retrain our Q-Function model combining the data of all previous blocks and run the updated model and its derived policy in the subsequent block. We always start the first block, in absence of any data, with a random policy.
This process continues for all the blocks until the end, refining the models progressively. 

\section{Exploration Strategies and Metrics}

Our approach involves using an off the shelf product combined with a exploration strategy. We found that using our $\frac{1}{n}$ annealing schedule for $\epsilon$-greedy exploration provided us much better results over a fixed value( where n is the current iteration number). Our meta-learning approach was built on top of this baseline exploration strategy. We compare the performance of this approach against the Online Cover algorithm. We also showcase the results of a random A/B testing baseline to compare the performance gains with respect to the simplest approach. We used the Regret metric as mentioned in section \ref{Meta-Learning} to benchmark our meta learning approach.

\begin{figure}[h!]
  \centering
  \captionsetup{justification=centering}
  \begin{subfigure}[t]{0.8\linewidth}
    \includegraphics[width=\linewidth, height=3cm,valign=t]{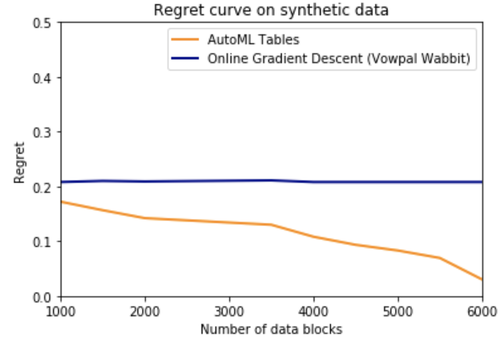}
    \caption{Results on the Synthetic dataset}
  \end{subfigure}
  \begin{subfigure}[t]{0.8\linewidth}
    \includegraphics[width=\linewidth, height=3cm,valign=t]{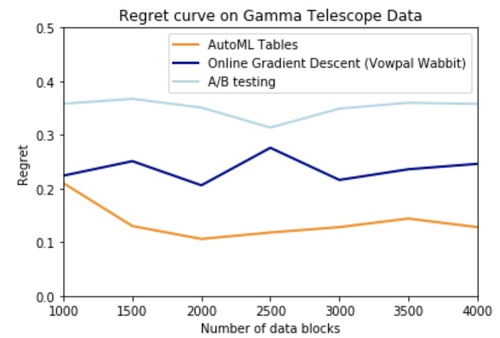}
    \caption{Results on the Gamma Telescope dataset}
  \end{subfigure}
  \begin{subfigure}[t]{0.8\linewidth}
    \includegraphics[width=\linewidth, height=3cm,valign=t]{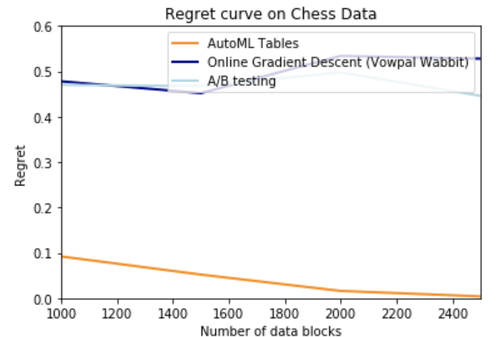}
    \caption{Results on the  Chess dataset}
  \end{subfigure}
  \begin{subfigure}[t]{0.8\linewidth}
    \includegraphics[width=\linewidth, height=3cm,valign=t]{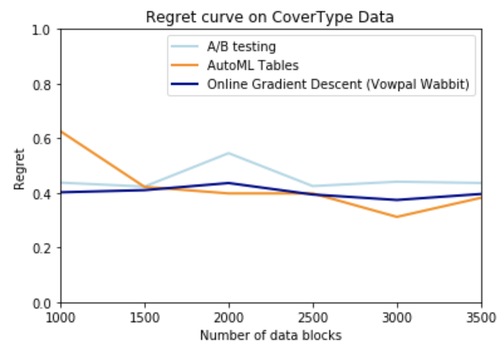}
    \caption{Results on the CoverType dataset}
  \end{subfigure}
  \begin{subfigure}[t]{0.8\linewidth}
    \includegraphics[width=\linewidth, height=3cm,valign=t]{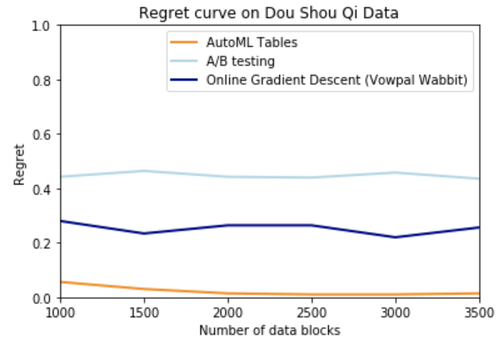}
    \caption{Results on the Dou Shou Qi dataset}
  \end{subfigure}
    \caption{Comparison of our model performance with the online gradient descent in Vowpal Wabbit as well as simple A/B testing. We are computing regret which is a performance measure for how well these algorithms perform optimally}
  \label{fig:Performance_comparisons}
\end{figure}

\section{Pipeline}

Our data collection approach has been explained in Section \ref{Public Dataset}. We sample these batch of experiences guided by our exploration strategy.

We applied the necessary pre-processing steps via Google Cloud BigQuery. This also dynamically handles data splits. These are then fed into the input pipeline for the AutoML Tables API. We maintained a held out inference set to verify the performance on unseen data. To verify our approach we ran multiple runs of the same parameter configurations and  experimented with varying number of training samples, standard noise, factor variance. With increasing training samples, we would expect our model to better map the relationship between the state and action space and perform more accurately. Also the expected value of increasing factor size, and  number of action of the Mean Absolute Error should be decreased. 

We used the online cover algorithm  implementation for Contextual Bandit in the Vowpal Wabbit library \citep{Vowpalwabbit}\citep{DBLP:journals/corr/AgarwalHKLLS14} and compared performance of both on the synthetic data. Finally to standardize our model's performance, we deployed our model in the supervised learning datasets proposed in \citep{a-contextual-bandit-bake-off} and compared its performance against other state of the art models. While the Online Cover algorithm has shown promising results and ease of reproducibility with the open source implementation, \citep{riquelme2018deep} provided a survey of multiple algorithms which can also be used to benchmark our model performance.
For this dataset, the $\epsilon$-greedy function had it's Epsilon value anneal over time from 0.9 to eventually 0.0. This is reflected in its decreased average regret.

The $\epsilon$-greedy approach achieves a balance between exploration and exploitation, as described more abstractly in the introduction.

\section{ Experimental Results}\label{Experimental Results}

Figure ~\ref{fig:Performance_comparisons} showcases the performance of our  bandit model powered by AutoML Tables. It performs well and sometimes exceptionally better than previous work on the different datasets. It's worth noting that this low regret prompted us to suspect data leakage, but that turned out not to be the case when we inspected the feature importance and found no proxy for the classification. The results shown have been averaged over multiple runs with the datasets being randomly reshuffled in each run. We experimented with some of the hyperparameters with AutoML handling most of these operations in under the hood as mentioned in section \ref{Meta-Learning}

\section{Conclusion and Next Steps}

We propose the use of an off the shelf meta-learning approach to solve the Contextual Bandits problem with no custom feature engineering required. Our internally generated synthetic environment allowed us to quickly iterate and experiment with different environment conditions and policies. We showcase competitive results on various public datasets, converging to low regret quickly compared to the Online Cover algorithm. 

We have showcased our meta learning model guided by a given e-greedy policy. As mentioned, our approach is agnostic to the exploration strategy used and future work would involve  experimenting with strategies such as UCB, Thompson sampling, bootstrapping models etc.
An interesting result would be on the ability of our meta-learning approach to adapt to time dependent environments. Incrementally adding a noise parameter to the environment would be a telling experiment to mimic real world scenarios.

\bibliographystyle{ACM-Reference-Format}
\bibliography{arxiv_paper_bib}


\begin{thebibliography}{00}

%%% the Web address.

\ifx \showCODEN    \undefined \def \showCODEN     #1{\unskip}     \fi
\ifx \showDOI      \undefined \def \showDOI       #1{#1}\fi
\ifx \showISBNx    \undefined \def \showISBNx     #1{\unskip}     \fi
\ifx \showISBNxiii \undefined \def \showISBNxiii  #1{\unskip}     \fi
\ifx \showISSN     \undefined \def \showISSN      #1{\unskip}     \fi
\ifx \showLCCN     \undefined \def \showLCCN      #1{\unskip}     \fi
\ifx \shownote     \undefined \def \shownote      #1{#1}          \fi
\ifx \showarticletitle \undefined \def \showarticletitle #1{#1}   \fi
\ifx \showURL      \undefined \def \showURL       {\relax}        \fi
% invisible to TeX
\providecommand\bibfield[2]{#2}
\providecommand\bibinfo[2]{#2}
\providecommand\natexlab[1]{#1}
\providecommand\showeprint[2][]{arXiv:#2}

\bibitem[\protect\citeauthoryear{Agarwal, Hsu, Kale, Langford, Li, and
  Schapire}{Agarwal et~al\mbox{.}}{2014}]%
        {DBLP:journals/corr/AgarwalHKLLS14}
\bibfield{author}{\bibinfo{person}{Alekh Agarwal}, \bibinfo{person}{Daniel~J.
  Hsu}, \bibinfo{person}{Satyen Kale}, \bibinfo{person}{John Langford},
  \bibinfo{person}{Lihong Li}, {and} \bibinfo{person}{Robert~E. Schapire}.}
  \bibinfo{year}{2014}\natexlab{}.
\newblock \showarticletitle{Taming the Monster: {A} Fast and Simple Algorithm
  for Contextual Bandits}.
\newblock \bibinfo{journal}{{\em CoRR\/}}  \bibinfo{volume}{abs/1402.0555}
  (\bibinfo{year}{2014}).
\newblock
\showeprint[arxiv]{1402.0555}
\showURL{%
\url{http://arxiv.org/abs/1402.0555}}


\bibitem[\protect\citeauthoryear{Bietti, Agarwal, and Langford}{Bietti
  et~al\mbox{.}}{2018}]%
        {a-contextual-bandit-bake-off}
\bibfield{author}{\bibinfo{person}{Alberto Bietti}, \bibinfo{person}{Alekh
  Agarwal}, {and} \bibinfo{person}{John Langford}.}
  \bibinfo{year}{2018}\natexlab{}.
\newblock \bibinfo{title}{A Contextual Bandit Bake-off}.
\newblock   (\bibinfo{date}{May} \bibinfo{year}{2018}).
\newblock
\showURL{%
\url{https://www.microsoft.com/en-us/research/publication/a-contextual-bandit-bake-off/}}


\bibitem[\protect\citeauthoryear{Blackard and Anderson}{Blackard and
  Anderson}{1998}]%
        {Covertype-Data-Set}
\bibfield{author}{\bibinfo{person}{Dean Blackard} {and}
  \bibinfo{person}{Anderson}.} \bibinfo{year}{1998}\natexlab{}.
\newblock \bibinfo{title}{{UCI} Machine Learning Repository}.
\newblock   (\bibinfo{year}{1998}).
\newblock
\showURL{%
\url{http://archive.ics.uci.edu/ml}}


\bibitem[\protect\citeauthoryear{Bock}{Bock}{2004}]%
        {Gamma-Telescope-Data-Set}
\bibfield{author}{\bibinfo{person}{Chilingarian A. Gaug M. Hakl F. Hengstebeck
  T. Jirina M. Klaschka J. Kotrc E. Savicky P. Towers S. Vaicilius A. Wittek~W.
  Bock, R.K.}} \bibinfo{year}{2004}\natexlab{}.
\newblock \bibinfo{title}{{UCI} Machine Learning Repository}.
\newblock   (\bibinfo{year}{2004}).
\newblock
\showURL{%
\url{http://archive.ics.uci.edu/ml}}


\bibitem[\protect\citeauthoryear{Dua and Graff}{Dua and Graff}{2017}]%
        {Dua:2019}
\bibfield{author}{\bibinfo{person}{Dheeru Dua} {and} \bibinfo{person}{Casey
  Graff}.} \bibinfo{year}{2017}\natexlab{}.
\newblock \bibinfo{title}{{UCI} Machine Learning Repository}.
\newblock   (\bibinfo{year}{2017}).
\newblock
\showURL{%
\url{http://archive.ics.uci.edu/ml}}


\bibitem[\protect\citeauthoryear{Elsken, Hendrik~Metzen, and Hutter}{Elsken
  et~al\mbox{.}}{2018}]%
        {NAS:Survey}
\bibfield{author}{\bibinfo{person}{Thomas Elsken}, \bibinfo{person}{Jan
  Hendrik~Metzen}, {and} \bibinfo{person}{Frank Hutter}.}
  \bibinfo{year}{2018}\natexlab{}.
\newblock \bibinfo{title}{Neural Architecture Search: A Survey}.
\newblock   (\bibinfo{date}{08} \bibinfo{year}{2018}).
\newblock


\bibitem[\protect\citeauthoryear{John~Langford and Strehl}{John~Langford and
  Strehl}{2007}]%
        {Vowpalwabbit}
\bibfield{author}{\bibinfo{person}{Lihong~Li John~Langford} {and}
  \bibinfo{person}{Alexander Strehl}.} \bibinfo{year}{2007}\natexlab{}.
\newblock \bibinfo{title}{{Vowpal wabbit open source project. In Technical
  Report, Yahoo!, 2007}}.
\newblock \bibinfo{howpublished}{\url{http://hunch.net/?p=309}}.
  (\bibinfo{date}{Dec.} \bibinfo{year}{2007}).
\newblock


\bibitem[\protect\citeauthoryear{Langford and Zhang}{Langford and
  Zhang}{2007}]%
        {Langford:2007:EAC:2981562.2981665}
\bibfield{author}{\bibinfo{person}{John Langford} {and} \bibinfo{person}{Tong
  Zhang}.} \bibinfo{year}{2007}\natexlab{}.
\newblock \showarticletitle{The Epoch-Greedy Algorithm for Contextual
  Multi-armed Bandits}. In \bibinfo{booktitle}{{\em Proceedings of the 20th
  International Conference on Neural Information Processing Systems}} {\em
  (\bibinfo{series}{NIPS'07})}. \bibinfo{publisher}{Curran Associates Inc.},
  \bibinfo{address}{USA}, \bibinfo{pages}{817--824}.
\newblock
\showISBNx{978-1-60560-352-0}
\showURL{%
\url{http://dl.acm.org/citation.cfm?id=2981562.2981665}}


\bibitem[\protect\citeauthoryear{Li, Chu, Langford, and Schapire}{Li
  et~al\mbox{.}}{2010}]%
        {DBLP:journals/corr/abs-1003-0146}
\bibfield{author}{\bibinfo{person}{Lihong Li}, \bibinfo{person}{Wei Chu},
  \bibinfo{person}{John Langford}, {and} \bibinfo{person}{Robert~E. Schapire}.}
  \bibinfo{year}{2010}\natexlab{}.
\newblock \showarticletitle{A Contextual-Bandit Approach to Personalized News
  Article Recommendation}.
\newblock \bibinfo{journal}{{\em CoRR\/}}  \bibinfo{volume}{abs/1003.0146}
  (\bibinfo{year}{2010}).
\newblock
\showeprint[arxiv]{1003.0146}
\showURL{%
\url{http://arxiv.org/abs/1003.0146}}


\bibitem[\protect\citeauthoryear{Lu}{Lu}{2019}]%
        {GoogleAIBlog}
\bibfield{author}{\bibinfo{person}{Yifeng Lu}.}
  \bibinfo{year}{2019}\natexlab{}.
\newblock \bibinfo{title}{{An End-to-End AutoML Solution for Tabular Data at
  KaggleDays}}.
\newblock
  \bibinfo{howpublished}{\url{http://ai.googleblog.com/2019/05/an-end-to-end-automl-solution-for.html}}.
    (\bibinfo{date}{May} \bibinfo{year}{2019}).
\newblock


\bibitem[\protect\citeauthoryear{Riquelme, Tucker, and Snoek}{Riquelme
  et~al\mbox{.}}{2018}]%
        {riquelme2018deep}
\bibfield{author}{\bibinfo{person}{Carlos Riquelme}, \bibinfo{person}{George
  Tucker}, {and} \bibinfo{person}{Jasper Snoek}.}
  \bibinfo{year}{2018}\natexlab{}.
\newblock \showarticletitle{Deep Bayesian Bandits Showdown: An Empirical
  Comparison of Bayesian Deep Networks for Thompson Sampling}. In
  \bibinfo{booktitle}{{\em International Conference on Learning
  Representations}}.
\newblock
\showURL{%
\url{https://openreview.net/forum?id=SyYe6k-CW}}


\bibitem[\protect\citeauthoryear{Sawant, Namballa, Sadagopan, and
  Nassif}{Sawant et~al\mbox{.}}{2018}]%
        {DBLP:journals/corr/abs-1810-01859}
\bibfield{author}{\bibinfo{person}{Neela Sawant}, \bibinfo{person}{Chitti~Babu
  Namballa}, \bibinfo{person}{Narayanan Sadagopan}, {and}
  \bibinfo{person}{Houssam Nassif}.} \bibinfo{year}{2018}\natexlab{}.
\newblock \showarticletitle{Contextual Multi-Armed Bandits for Causal
  Marketing}.
\newblock \bibinfo{journal}{{\em CoRR\/}}  \bibinfo{volume}{abs/1810.01859}
  (\bibinfo{year}{2018}).
\newblock
\showeprint[arxiv]{1810.01859}
\showURL{%
\url{http://arxiv.org/abs/1810.01859}}


\bibitem[\protect\citeauthoryear{Sharaf and III}{Sharaf and III}{2019}]%
        {DBLP:journals/corr/abs-1901-08159}
\bibfield{author}{\bibinfo{person}{Amr Sharaf} {and}
  \bibinfo{person}{Hal~Daum{\'{e}} III}.} \bibinfo{year}{2019}\natexlab{}.
\newblock \showarticletitle{Meta-Learning for Contextual Bandit Exploration}.
\newblock \bibinfo{journal}{{\em CoRR\/}}  \bibinfo{volume}{abs/1901.08159}
  (\bibinfo{year}{2019}).
\newblock
\showeprint[arxiv]{1901.08159}
\showURL{%
\url{http://arxiv.org/abs/1901.08159}}


\bibitem[\protect\citeauthoryear{Slivkins}{Slivkins}{2019}]%
        {DBLP:journals/corr/abs-1904-07272}
\bibfield{author}{\bibinfo{person}{Aleksandrs Slivkins}.}
  \bibinfo{year}{2019}\natexlab{}.
\newblock \showarticletitle{Introduction to Multi-Armed Bandits}.
\newblock \bibinfo{journal}{{\em CoRR\/}}  \bibinfo{volume}{abs/1904.07272}
  (\bibinfo{year}{2019}).
\newblock
\showeprint[arxiv]{1904.07272}
\showURL{%
\url{http://arxiv.org/abs/1904.07272}}


\bibitem[\protect\citeauthoryear{van Rijn and Vis}{van Rijn and Vis}{2016}]%
        {Dou-Shou-Qi}
\bibfield{author}{\bibinfo{person}{J.~N. van Rijn} {and} \bibinfo{person}{J.~K.
  Vis}.} \bibinfo{year}{2016}\natexlab{}.
\newblock \bibinfo{title}{{UCI} Machine Learning Repository}.
\newblock   (\bibinfo{year}{2016}).
\newblock
\showURL{%
\url{https://arxiv.org/abs/1604.07312}}


\bibitem[\protect\citeauthoryear{Zoph and Le}{Zoph and Le}{2016}]%
        {DBLP:journals/corr/ZophL16}
\bibfield{author}{\bibinfo{person}{Barret Zoph} {and} \bibinfo{person}{Quoc~V.
  Le}.} \bibinfo{year}{2016}\natexlab{}.
\newblock \showarticletitle{Neural Architecture Search with Reinforcement
  Learning}.
\newblock \bibinfo{journal}{{\em CoRR\/}}  \bibinfo{volume}{abs/1611.01578}
  (\bibinfo{year}{2016}).
\newblock
\showeprint[arxiv]{1611.01578}
\showURL{%
\url{http://arxiv.org/abs/1611.01578}}


\end{thebibliography}

\end{document}